\documentclass[10pt, a4paper]{article}
\usepackage{comment}
\usepackage[]{lrec2026} 
\usepackage{xcolor}
\usepackage{amsmath}
\usepackage{multirow}

\title{{DeID-Clinic}: A Risk-Aware Pseudonymization Framework for Clinical Text De-identification and Re-identification Risk Assessment
} 

\name{
Angel Paul$^{1,\dagger}$, Dhivin Shaji$^{1,\dagger}$, Lifeng Han$^{2,3,*}$ \\ {\large \textbf{Warren Del-Pinto}$^1$, \textbf{Goran Nenadic}$^1$, \textbf{Suzan Verberne}$^2$}
}

\address{ $^1$University of Manchester, Manchester, UK \\ $^2$Leiden Institute of Advanced Computer Science (LIACS), Leiden University, NL\\
$^3$Biomedical Data Sciences,
Leiden University Medical Center, Leiden, NL\\ 
         $^{\dagger}$co-first \textit{$^*$corresponding}: \{l.han, s.verberne\}@liacs.leidenuniv.nl \\}

\abstract{
The increasing availability of sensitive textual data has created an urgent need for robust de-identification methods that enable compliant data sharing while preserving downstream utility. This paper presents DeID-Clinic, a multi-layered framework for automated pseudonymization and re-identification risk assessment of clinical free-text data. Our approach integrates domain-adapted transformer models, including BioBERT and ClinicalBERT, into the MASK de-identification framework to improve the detection and masking of protected health information (PHI). Beyond entity recognition, we introduce a novel document-level risk assessment module that quantifies residual re-identification risk using a combination of k-anonymity, l-diversity, t-closeness, contextual similarity, and entity co-occurrence analysis. Experiments conducted on the i2b2 2014 de-identification dataset demonstrate strong performance, achieving macro-level F1 scores above 0.96 for several entity categories, while enabling quantitative prioritization of high-risk documents for further review. Our results highlight the effectiveness of combining neural de-identification with explicit risk modeling, supporting privacy-preserving data sharing in sensitive domains. Although evaluated on clinical text, the proposed framework is generalizable to other privacy-critical domains such as legal and administrative documents, where reliable pseudonymization and risk-aware anonymization are essential.
 \\ \newline  }

\begin{document}

\maketitleabstract

\section{Introduction}
\label{sec:intro}
The widespread adoption of electronic health records and other sensitive textual datasets has created an urgent need for reliable de-identification methods that not only remove personally identifiable information but also quantify the residual risk of re-identification \cite{scaiano2016unified,subramanian2024patient,sarkar2024identification}. While recent neural approaches have achieved high accuracy in detecting protected health information, most existing systems focus solely on entity masking without assessing whether the resulting text remains vulnerable to re-identification through contextual clues or rare entity combinations \cite{sondeck2025practical}. This limitation poses a significant challenge for privacy-preserving data sharing, as effective anonymization requires both accurate pseudonymization and rigorous risk evaluation. Addressing this gap, we propose a \textit{risk-aware de-identification} framework that integrates transformer-based entity recognition with document-level privacy risk assessment, enabling more reliable and accountable anonymization of clinical free-text data.

The need for privacy-preserving text processing is particularly critical in healthcare, where clinical narratives contain sensitive patient information protected under regulations such as GDPR and HIPAA \cite{ElEmam2006,Voigt2017,edemekong2024health}. De-identification aims to reduce the risk of re-identification by removing or replacing sensitive information while preserving data utility for research and clinical applications \cite{Sweeney2002a,Dankar2012}.
For example, a clinical sentence containing a patient name, date, and location may be transformed into a pseudonymized version that maintains clinical meaning but protects individual privacy \cite{Meystre2010,Stubbs2015}. 

Recent advances in neural language models, particularly transformer-based architectures such as BERT and its domain-specific variants, have significantly improved the accuracy of identifying sensitive entities in text. Models such as BioBERT and ClinicalBERT \cite{Lee2020,Alsentzer2019} leverage domain-specific pre-training to better capture the linguistic characteristics of clinical narratives. These models have demonstrated strong performance in named entity recognition tasks, making them promising candidates for automated de-identification.
However, accurate entity detection alone does not guarantee effective privacy protection \cite{Kovacevic2024}. Even after pseudonymization, residual information such as rare entity combinations or unique contextual patterns may enable re-identification. Consequently, there is a growing need for methods that not only perform de-identification but also quantify the residual risk associated with anonymized text. Such risk-aware approaches are critical for supporting responsible data sharing and ensuring compliance with privacy regulations \cite{Hara2018}.

%
To address these challenges, we present \textbf{DeID-Clinic}, a multi-layered framework for automated pseudonymization and re-identification risk assessment of clinical free-text data. Our approach integrates domain-adapted transformer models into the open-sourced MASK framework \cite{milosevic2020mask} to improve entity detection and masking and introduces a document-level risk assessment module to quantify residual privacy risks \footnote{this is an extended work from our 2-page poster paper \cite{shaji2025identifying}. In this longer paper, we describe the details on methodology design and carry out more experimental evaluations and analysis.}.

This work advances privacy-preserving language processing by introducing a risk-aware pseudonymization framework that integrates neural entity recognition with quantitative privacy risk estimation.
The key contributions are:
1) Risk-aware pseudonymization framework:
We propose DeID-Clinic, a unified framework that combines neural de-identification and document-level re-identification risk assessment, enabling both automated pseudonymization and quantitative privacy evaluation.
2) Document-level privacy risk modeling:
We introduce a novel risk scoring method that integrates classical anonymization metrics (k-anonymity, l-diversity, t-closeness) with contextual embedding similarity and entity co-occurrence analysis to estimate residual re-identification risk in free-text documents.
3) Integration of domain-adapted transformer models:
We extend the MASK platform by incorporating BioBERT and ClinicalBERT models, improving sensitive entity detection accuracy on clinical text.
4) Comprehensive experimental and case-study evaluation:
We evaluate the framework on the i2b2 2014 dataset and demonstrate its effectiveness in both entity detection performance and risk-aware document prioritization.
\section{Related Work}
Automated de-identification of clinical text has been extensively studied, with approaches evolving from rule-based systems to modern deep learning models. In addition, emerging research has begun to explore methods for assessing re-identification risk after de-identification.

\subsection{Rule-based and Traditional ML
}

Early de-identification systems primarily relied on rule-based approaches, which use manually defined patterns and dictionaries to identify sensitive entities such as names, dates, and locations \cite{friedlin2008software,Meystre2014}. While effective in structured settings, rule-based systems often lack flexibility and struggle with linguistic variability and ambiguity in clinical narratives.

Machine learning approaches such as Conditional Random Fields (CRFs) were later introduced, allowing models to learn entity patterns directly from annotated data \cite{Yang2019,Liu2017}. These methods improved adaptability and performance but still faced limitations in capturing long-range dependencies and contextual relationships.

Recurrent neural network architectures, particularly BiLSTM models, further improved performance by modeling sequential dependencies in text \cite{Dernoncourt2017,Kim2018}. However, these models often require extensive feature engineering and may struggle with complex contextual interactions, e.g., in long clinical documents \cite{Lin2020}.

\subsection{Transformer-based
De-identification
}

The introduction of transformer-based language models has significantly advanced clinical de-identification. Domain-adapted models such as BioBERT \cite{Lee2020} and ClinicalBERT \cite{Alsentzer2019} leverage pre-training on biomedical and clinical corpora to improve entity recognition performance. These models have demonstrated strong results across multiple clinical NLP tasks, including PHI detection.
Recent systems have successfully applied transformer-based architectures to de-identification tasks, achieving state-of-the-art performance while reducing the need for manual feature engineering \cite{Kraljevic2023}. 
\subsection{De-identification Frameworks and Systems}

The MASK framework \cite{milosevic2020mask} provides a flexible open-sourced platform for clinical text de-identification, supporting multiple named entity recognition models and masking strategies \footnote{\url{https://github.com/icescentral/MASK_public}}. MASK enables both redaction and pseudonymization and allows integration of custom NER models. Its modular design makes it suitable for deployment in real-world clinical environments.

Another widely used system is Philter, a rule-based de-identification tool designed for large-scale clinical text processing \cite{Hartman2020}. Philter offers high customizability and transparency through manually defined filtering rules, making it particularly suitable for environments where explainability and precise control are required. However, rule-based approaches may require extensive manual tuning and may not generalize well across datasets.

More recently, AnonCAT, integrated within the MedCAT ecosystem, combines transformer-based models with biomedical knowledge graphs to improve de-identification accuracy and contextual understanding \cite{Vakili2022,Kraljevic2023}. By leveraging domain knowledge and fine-tuning strategies, AnonCAT provides a flexible and scalable solution for clinical text anonymization.
\subsection{Risk Assessment and Privacy Evaluation}
While significant progress has been made in entity detection and masking, fewer studies have focused on evaluating residual re-identification risk after de-identification. Privacy models such as k-anonymity \cite{Sweeney2002b}, l-diversity \cite{Machanavajjhala2007}, and t-closeness \cite{Li2007} provide formal mechanisms for assessing identifiability in structured data.

These methods have been adapted to evaluate privacy risks in clinical datasets \cite{Dankar2012,Hara2018}. However, their integration into automated de-identification pipelines for unstructured clinical text remains limited.

In this work, we extend existing de-identification frameworks by incorporating document-level risk assessment alongside neural pseudonymization, enabling both sensitive entity detection and quantitative evaluation of residual privacy risk.

\begin{figure*}[t!]
\centering
\includegraphics[width=0.95\textwidth]{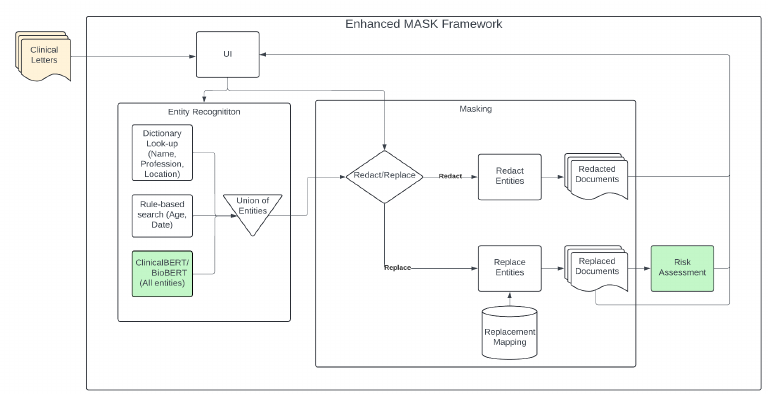}
    \caption{Detailed Architecture of Enhanced MASK framework - {DeID-Clinic}}
    \label{fig:workflow-diag}
\end{figure*}
\section{Methods and Design}




\subsection{Architecture Overview}

The system architecture, as depicted in the diagram (Figure \ref{fig:workflow-diag}), is divided into three major sections: Entity Recognition, Masking, and Risk Assessment. The clinical letters serve as the input to the system, and they are processed through multiple pipelines before final redacted or replaced documents are produced.
The steps are as follows:
1) \textbf{Data Ingestion}: Clinical letters are fed into the system via the user interface (UI), allowing users to upload documents in bulk for de-identification. 
2) \textbf{Entity Recognition}: We applied several techniques to identify sensitive information in the clinical letters, such as names, professions, dates, ages, and locations. To accommodate the complexity of clinical data, we integrated multiple approaches:
Dictionary look-up leverages predefined dictionaries to detect common sensitive information categories such as names, professions, and locations, ensuring consistent identification of widely known terms across the dataset. Rule-based search handles entities with structured formats, such as dates and ages, using regular expressions to capture variations in formatting. Additionally, a machine learning model based on pre-trained BioBERT and ClinicalBERT is integrated to enhance entity recognition accuracy by capturing contextual information beyond the capabilities of rule-based and dictionary methods.
3) \textbf{Union of Entities}: After we have applied the various methods of entity recognition, the system combines the identified entities into a unified list for further processing, ensuring comprehensive entity recognition. This phase 
incorporates multiple techniques that complement each other.
4) \textbf{Masking Strategies}: The next phase involves applying masking strategies. Users can choose between Redaction and Replacement. Redaction involves replacing the identified entities with placeholders (e.g., "XXX-Name"). Replacement, on the other hand, substitutes sensitive information with synthetically generated or random replacements to maintain the structure of the original document. Replacement is more suitable for contexts where document coherence needs to be preserved, such as for clinical research purposes \cite{Neamatullah2008}. 
The Replacement mapping is stored and later used for reference in the risk assessment stage. 
5) \textbf{Risk Assessment}: The probability of re-identification of the de-identified data is evaluated, ensuring that the transformation process sufficiently anonymizes sensitive information \cite{ElEmam2008}. 
The integration of these metrics enables a quantitative assessment of the risk associated with the data after de-identification.
6) \textbf{Final Output}: Depending on the chosen masking strategy, the system generates either redacted or replaced documents. These documents are returned to the user via the UI, alongside a risk assessment report.

By integrating advanced language models with robust masking strategies and risk assessment techniques, 
this architecture enables unified pseudonymization and quantitative risk assessment within a single processing pipeline, supporting privacy-aware text release workflows.


\subsection{The Risk Assessment Framework}
\label{subsec:risk-assess}
We implement a set of risk metrics 
to evaluate the robustness of the de-identification process and mitigate the risk of re-identification. 

For entity extraction with context, each entity is paired with a window of surrounding words to form a quasi-identifier, simulating realistic re-identification scenarios where attackers may have access to auxiliary information. Text is tokenized using the BertTokenizerFast, and character indices are aligned with token indices to accurately extract contextual spans. 

For k-anonymity \cite{Sweeney2002b}, entities are grouped according to their quasi-identifiers, defined by both entity type and contextual information. Context is represented through high-dimensional contextual embeddings that encode semantic meaning. The number of similar records within each group determines the anonymity level, and the smallest group size defines the overall k-anonymity score. 

L-diversity \cite{Machanavajjhala2007} is computed by measuring the number of distinct sensitive attribute values (e.g., age or profession) within each quasi-identifier group to ensure sufficient diversity. 

Unicity \cite{de2013unique} is measured by counting unique quasi-identifier combinations, where higher uniqueness implies increased re-identification risk. 

The quasi-identifier risk likelihood is estimated by assigning each combination a probability inversely proportional to its dataset frequency, and averaging these probabilities to approximate expected re-identification risk \cite{el2011systematic}. 

T-closeness \cite{Li2007} is evaluated by comparing the distribution of sensitive attributes within each group to the global distribution across the dataset using Kullback--Leibler divergence; smaller divergence indicates stronger privacy preservation. Cosine similarity is used to measure contextual similarity between embeddings \cite{yu2022cosbert}. Similar contexts across documents (scores near 1) indicate common entities with lower risk, whereas low similarity (scores near 0) suggests uniqueness and higher potential re-identification risk. 

After computing these metrics for both the original and de-identified datasets, a comparative risk assessment is conducted. The framework analyzes entity frequencies and co-occurrences, as combinations of entities increase identifiability. Each co-occurrence is assigned a sensitivity weight, and the document-level risk score (RS) is defined as:

\begin{equation}
\text{RS} = \sum(\text{EF} + \text{CoWeight})
\end{equation}

where EF denotes entity frequency and CoWeight denotes co-occurrence weight. The system further counts contexts with cosine similarity below a threshold of 0.5, and computes the proportion of unique contexts within each document relative to all contexts. This proportion is combined with the co-occurrence statistics to obtain the final risk score (FRS):

\begin{equation}
\text{FRS} =
\sum(\text{EF} + \text{CoWeight}) \times
\left(\frac{\text{Count}}{\text{TotalCount}}\right) \times 100
\end{equation}

The resulting score is expressed as a percentage and used to assign documents to three risk categories: low risk (below 25\%), moderate risk (25--50\%), and high risk (above 50\%). Finally, documents are prioritized accordingly, with high-risk documents requiring manual review and stricter de-identification, while low-risk documents require minimal intervention.

Unlike traditional anonymization approaches that focus solely on entity removal, our risk assessment framework explicitly models residual identifiability after pseudonymization. By combining statistical privacy metrics with contextual semantic similarity, the proposed approach provides a practical approximation of real-world re-identification risk, where adversaries may exploit contextual clues rather than isolated identifiers. This enables more informed decisions regarding data release and manual review prioritization.
\section{Experimental Evaluation}
\subsection{Dataset Overview}

The dataset used in this work is the i2b2/UTHealth De-identification and Heart Disease Risk Factors dataset, specifically the 2014 PHI Gold Set 1 and 2 \cite{Stubbs2014}, which is part of the National NLP Clinical Challenges (n2c2) initiative (n2c2 NLP Research Data Sets).\footnote{\url{https://n2c2.dbmi.hms.harvard.edu}} This dataset comprises de-identified clinical notes that are extensively annotated for Protected Health Information (PHI) and are intended for evaluating and advancing the performance of de-identification systems in clinical settings.
The dataset is sourced from the Research Patient Data Registry (RPDR) at Partners Healthcare and was manually annotated by domain experts. The dataset consists of 790 clinical notes spanning multiple years of patient data. 
The dataset contains the following de-identification entities: 4,456 names, 7,495 dates, 897 medical identifiers, 2,767 locations, 234 professions, 323 contact details, and 1,424 ages.
This annotation process provides a dataset for training and evaluating de-identification models across a diverse range of PHI categories.
The dataset is structured to facilitate research in clinical text de-identification, with annotations corresponding to several categories of PHI. 
The entity categories 
can be further categorised as direct identifiers (e.g., names and contact information) and quasi-identifiers (e.g., ages, locations). Direct identifiers refer to information that identifies an individual, such as names, contact details, or Social Security numbers. Quasi-identifiers are pieces of information that do not directly identify an individual but can be combined with other data to re-identify someone \cite{scaiano2016unified}.

\subsection{Model Setup and Finetuning}
The BioBERT and ClinicalBERT models are integrated into the MASK framework to identify and classify sensitive entities in clinical text. The models are further fine-tuned using the i2b2 dataset to adapt them for clinical NER tasks.


We implement NER following a common token classification approach, where each token in a sequence is assigned a label. Given the nature of clinical notes, where a single entity may span multiple tokens, the model uses \textbf{BIO} tagging (Begin, Inside, Outside tagging) to ensure that multi-token entities are labelled correctly. For the sentence, ``John Smith visited the hospital on 12th August 2024.", the BIO tags might be as in Table \ref{tab:bio-tagging}. 
\begin{table*}[h]
\centering
\begin{tabularx}{\textwidth}{|l|X|X|l|l|l|l|X|l|X|}
\hline
\textbf{Token} & John & Smith & visited & the & hospital & on & 12th & August & 2024 \\
\hline
\textbf{Tag} & B-NAME & I-NAME & O & O & B-ORG & O & B-DATE & I-DATE & I-DATE \\
\hline
\end{tabularx}
\caption{BIO Tagging}
\label{tab:bio-tagging}
\end{table*}

Here, the name ``John Smith" is recognised as a ``NAME" entity, the ``hospital" as an ``ORG" (Organization) entity, and ``12th August 2024" as a ``DATE" entity. Each of these entities has appropriate "B" and "I" tags depending on whether the token is at the beginning (B) or inside (I) of the entity.

The finetuning of Bio/ClinicalBERT involves the following key steps:
I) {Data Preprocessing}: 
1) The input text is split into sentences using the \texttt{sent\_tokenize} function from NLTK, ensuring that the model processes manageable text chunks.
2) Each sentence is then tokenized using the BioBERT and ClinicalBERT tokenizer, which handles subword tokens. This is crucial for preserving the granularity of clinical entities.
3) The tokenized sentences are padded to a maximum length of 75 tokens, ensuring uniformity in batch processing.

II) {Training Setup}:
1) The model is set up to utilise a GPU (T4) with CUDA when available; otherwise, it will default to running on the CPU. 
2) BioBERT and ClinicalBERT are finetuned on the i2b2 2014 dataset, specifically focusing on the NER task.
3) The training process uses the AdamW optimizer with a learning rate of $3\times 10^{-5}$ and weight decay to prevent overfitting. This value was found to be small enough to ensure stable convergence during training while allowing the model to learn efficiently from the dataset.
4) The model was trained using a batch size of 4 to optimize memory utilisation on GPU hardware. The training process was carried out over 20 epochs, with loss computed at each step.

III)
{Learning and Evaluation}:
1) During the learning process, the model's parameters were updated iteratively based on the cross-entropy loss between predicted and true labels.
2) After each epoch, the model's performance was evaluated using validation data, and loss curves were plotted to monitor overfitting or underfitting tendencies.
3) The primary evaluation metrics used were Precision, Recall, F1-score, and Accuracy.

\subsection{BioBERT and ClinicalBERT Results}

\begin{table*}[t]
\centering
\begin{tabular}{|l|ccc|ccc|r|}
\hline
\multirow{2}{*}{\textbf{Entity}} & \multicolumn{3}{c|}{\textbf{BioBERT}} & \multicolumn{3}{c|}{\textbf{ClinicalBERT}} & \multirow{2}{*}{\textbf{ items}} \\
\cline{2-7}
 & \textbf{P} & \textbf{R} & \textbf{F1} & \textbf{P} & \textbf{R} & \textbf{F1} & \\
\hline
NAME        & 0.963 & 0.960 & \textbf{0.985} & \textbf{0.970} & \textbf{0.970} & 0.970 & 622 \\
DATE        & \textbf{0.974} & 0.974 & 0.974 & 0.960 & \textbf{0.982} & \textbf{0.971} & 655 \\
ID          & 0.948 & 0.973 & 0.988 & \textbf{0.986} & \textbf{0.993} & \textbf{0.989} & 75 \\
AGE         & \textbf{0.956} & \textbf{0.978} & \textbf{0.967} & 0.930 & 0.974 & 0.951 & 89 \\
LOCATION    & \textbf{0.933} & 0.903 & \textbf{0.928} & 0.919 & \textbf{0.922} & 0.921 & 278 \\
CONTACT     & \textbf{0.955} & 0.928 & 0.941 & 0.947 & \textbf{0.960} & \textbf{0.954} & 69 \\
PROFESSION  & 0.810 & 0.630 & \textbf{0.748} & \textbf{0.844} & \textbf{0.643} & 0.730 & 27 \\
\hline
\textbf{Micro Avg}  & \textbf{0.959} & \textbf{0.964} & \textbf{0.965} & 0.955 & 0.963 & 0.959 & 1816 \\
\textbf{Macro Avg}  & 0.817 & 0.793 & 0.804 & \textbf{0.820} & \textbf{0.805} & \textbf{0.811} & 1816 \\
\textbf{Weighted Avg} & \textbf{0.958} & \textbf{0.961} & \textbf{0.964} & 0.953 & 0.963 & 0.958 & 1816 \\
\hline
\end{tabular}
\caption{Evaluations of BioBERT and ClinicalBERT models with higher scores in bold}
\label{tab:merged-classification-report-bio-n-clinical-BERT}
\end{table*}

Based on the comparisons in Table \ref{tab:merged-classification-report-bio-n-clinical-BERT} of BioBERT and ClinicalBERT in each entity category, we summarise the results as follows.
1), BioBERT wins more precision than ClinicalBERT, e.g. on DATE, AGE, LOCATION, CONTACT. 
2), ClinicalBERT wins more recall than BioBERT, e.g. on NAME, DATE, ID, LOCATION, CONTACT, and PROFESSION, except for the only entity category AGE.
3), BioBERT wins four entities on F1, i.e. NAME, AGE, LOCATION, and PROFESSION.
4), ClinicalBERT wins the rest three entity types, i.e. DATE, ID, and CONTACT.
5), in average across all entities, BioBERT wins micro avg scores and weighted avg scores, while ClinicalBERT wins macro avg scores on P/R/F1.

The F1 scores across all entity types mostly fall between 0.92 (LOCATION) and 0.99 (ID), except for PROFESSION who has the lowest F1 scores 0.748 (BioBERT) and 0.730 (ClinicalBERT).
%
%
This lower performance suggests that professions are more \textbf{ambiguous} and context-dependent, making them harder to identify in comparison to other entities such as IDs or Names; this is a known issue \cite{Uzuner2007,Dernoncourt2017}.


\subsection{Baseline Results} 

\begin{table*}[t]
\centering
\begin{tabular}{|l|ccc|ccc|ccc|}
\hline
\multirow{2}{*}{\textbf{Entity}} & \multicolumn{3}{c|}{\textbf{BERT}} & \multicolumn{3}{c|}{\textbf{BiLSTM}} & \multicolumn{3}{c|}{\textbf{CRF}} \\
\cline{2-10}
 & \textbf{P} & \textbf{R} & \textbf{F1} & \textbf{P} & \textbf{R} & \textbf{F1} & \textbf{P} & \textbf{R} & \textbf{F1} \\
\hline
NAME        & \textbf{0.979} & \textbf{0.980} & \textbf{0.979} & 0.980 & 0.960 & 0.970 & 0.940 & \textbf{0.980} & 0.960 \\
DATE        & 0.965 & \textbf{0.988} & \textbf{0.977} & 0.960 & 0.970 & 0.960 & \textbf{0.960} & {0.980} & {0.970} \\
ID          & 0.940 & {0.986} & 0.962 & \textbf{0.980} & 0.860 & 0.920 & 0.950 & \textbf{0.990} & \textbf{0.970} \\
AGE         & 0.878 & 0.908 & 0.893 & 0.740 & \textbf{0.990} & 0.850 & \textbf{0.910} & 0.980 & \textbf{0.940} \\
LOCATION    & \textbf{0.943} & 0.937 & \textbf{{0.940}} & 0.890 & {0.940} & 0.910 & 0.850 & \textbf{0.970} & 0.900 \\
CONTACT     & \textbf{0.953} & \textbf{0.994} & \textbf{0.973} & 0.950 & 0.930 & 0.940 & 0.930 & 0.990 & 0.960 \\
PROFESSION  & \textbf{0.907} & 0.680 & \textbf{0.777} & 0.810 & {0.780} & 0.790 & 0.470 & \textbf{0.930} & 0.630 \\
\hline
\textbf{Micro Avg}  & \textbf{0.962} & 0.972 & \textbf{0.967} & 0.940 & 0.950 & 0.950 & 0.920 & \textbf{0.980} & 0.950 \\
\textbf{Macro Avg}  & \textbf{0.821} & 0.809 & \textbf{0.813} & 0.790 & \textbf{0.800} & 0.790 & 0.750 & \textbf{0.850} & 0.790 \\
\textbf{Weighted Avg} & \textbf{0.960} & 0.972 & \textbf{0.966} & 0.950 & 0.950 & 0.950 & 0.930 & \textbf{0.980} & 0.950 \\
\hline
\end{tabular}
\caption{Comparison of evaluation metrics for BERT, BiLSTM, and CRF models.}
\label{tab:merged-classification-report-bert-lstm-crf}
\end{table*}

We also list the comparisons on BERT, BiLSTM, and CRF in Table \ref{tab:merged-classification-report-bert-lstm-crf}, where it shows that the BERT model wins the most Precision scores on 4 entities, versus BiLSTM (1) and CRF (2). In comparison, the CRF model 
wins the most Recall scores on 4 entities (NAME, ID, LOCATION, PROFESSION), versus BERT (3 including 1 tie) and BiLSTM (1), which indicates that CRFs produce more false positives for the sake of true positives. 
This is especially true for the PROFESSION entity type, where CRFs give the lowest precision score of 0.470. In contrast, the BERT model has much higher Precision than Recall (0.907 vs 0.680), indicating that it sacrifices potential true outputs by restricting false positives. Interestingly, the BiLSTM model has a more balanced P/R on the PROFESSION category (0.81 vs 0.78). 
Looking at both Table \ref{tab:merged-classification-report-bio-n-clinical-BERT} and \ref{tab:merged-classification-report-bert-lstm-crf}, we can see that the domain adapted models BioBERT and ClinicalBERT have improved the performance on entity types \textbf{NAME},  
\textbf{ID}, and \textbf{AGE} in comparison to BERT model from (0.979, 0.962, 0.893) to (0.985, 0.989, 0.967) on F1 scores. 

\begin{figure}[t]
\centering
\includegraphics[width=0.45\textwidth]{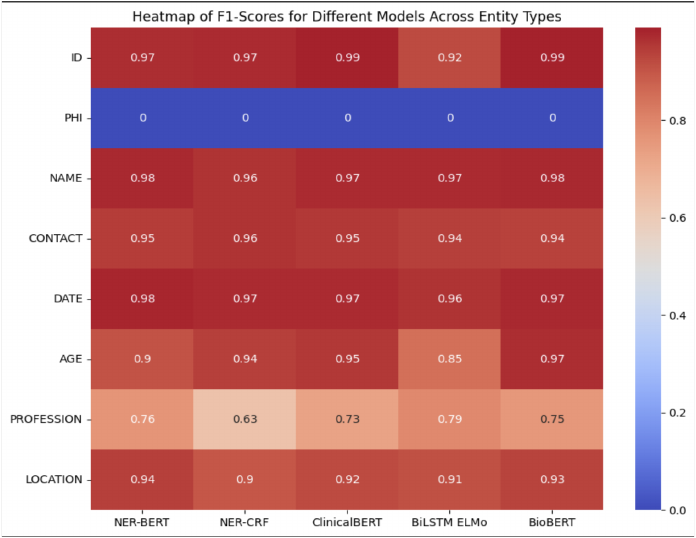}
    \caption{Heat-Map of All Evaluated Models}
    \label{fig:entity-level-heat-bioBERT-vs}
\end{figure}

\subsection{Heat-maps from All Models}

Figure \ref{fig:entity-level-heat-bioBERT-vs} presents the heat-map comparative results of F1-scores of MASK-BioBERT/ClinicalBERT and other MASK NER models. The results indicate that Biomed-Clinical BERT models consistently perform strongly or match the performance of other models in key entity categories, demonstrating their effectiveness for clinical data de-identification.
Notably, MASK-BioBERT demonstrated very high performance for \textbf{structured entity types} like IDs, Dates, and Names, where it consistently achieved higher or equal F1-scores compared to ClinicalBERT, NER-BERT, and NER-CRF, all likely benefiting from domain-specific pre-training. 

However, while MASK-BioBERT excels in structured entity recognition, its lower performance on context-dependent entities like \textbf{Profession} (0.75 F1-score) highlights its limitations in handling ambiguity. This is an area where even ClinicalBERT, which focuses on clinical texts, struggles. 

\subsection{Qualitative Case Study}
To illustrate the practical behavior of the proposed framework in realistic deployment scenarios, we conducted a qualitative case study on five randomly selected clinical documents from the i2b2 dataset. This case study complements the quantitative evaluation presented earlier by providing insight into system behavior at the document level, including entity detection accuracy and masking effectiveness.
These documents originally contained a total of 181 sensitive entities. 
From the annotations by two fluent English speakers (MSc graduates), the system using MASK-BioBERT and MASK-ClinicalBERT identified a total of 183 and 193 entities, respectively, in Table \ref{tab:ner-evaluation-comparison-M-bio-n-cliB} (\textit{left} and \textit{right}), with the key metrics observed from this test.


\begin{table}[t]
\centering
\setlength{\tabcolsep}{4pt}
\begin{tabular}{lcc}
\hline
Metric & M-BioBERT & M-ClinicalBERT \\
\hline
TP  & 171 & 181 \\
FP  & 10  & 12  \\
FN  & 2   & 0   \\
Precision & \textbf{0.9448} & 0.9378 \\
Recall    & 0.9948 & \textbf{1.0000} \\
F1        & 0.9648 & \textbf{0.9679} \\
\hline
\end{tabular}
\caption{NER evaluation comparisons.}
\label{tab:ner-evaluation-comparison-M-bio-n-cliB}
\end{table}

Here’s a detailed analysis of the metrics observed:
1) On Precision: The systems achieved the precision of 0.944 and 0.937, indicating a high level of accuracy in identifying entities. These values suggest that the majority of identified entities were correct, with only a small number of false positives identified when running the MASK-Bio/Clinical-BERT.
2) On Recall: A recall score of 0.9948 and 1 reflects the system's ability to correctly identify nearly all relevant entities present in the dataset. Out of all potential entities, only \textbf{2 false negatives} for MASK-BioBERT were missed, indicating a highly efficient model in terms of capturing the intended entities. In addition, MASK-ClinicalBERT had 0 false negatives on this task.
3) On F1 Score: The F1 score, calculated as the harmonic mean of precision and recall, stood at 0.964 and 0.967 for the two models. The robust F1 score illustrates that the model strikes an effective balance between precision and recall, providing reliable and consistent performance across different entity types.
The \textbf{10 and 12 false positives} from two systems indicate some over-identification by the models. These might arise from the model's sensitivity in identifying entities, where certain words are incorrectly flagged as entities, likely due to ambiguity in the context or overlaps in entity types. As seen in Table \ref{tab:sample-false-posi-bioBERT} ‘Clarkfield’ is identified as a name, when in reality it's actually a location.

\begin{table}[t]
\centering
\scriptsize
\setlength{\tabcolsep}{3pt}
\begin{tabular}{p{0.25\columnwidth} p{0.65\columnwidth}}
\hline
\textbf{Correct} &
\ttfamily <LOCATION id=P13 start=967 end=977 text=Clarkfield TYPE=HOSPITAL> \\
\hline
\textbf{False Positives} &
\ttfamily
2071(1576–1580, Date), US(1602–1604, Location), 2071(1745–1749, Date), Thiel(4026–4031, Name), 4(2138–2139, Age), Clarkfield(1317–1327, Name), Thiel(5890–5895, Name) \\
\hline
\end{tabular}
\caption{Sample false positives of MASK-BioBERT.}
\label{tab:sample-false-posi-bioBERT}
\end{table}

Despite these challenges, the system's high precision, recall, and F1 score suggest that it performs reliably in recognising sensitive information in clinical documents. These metrics highlight the system's potential to be a strong candidate for real-world applications in medical entity de-identification.

The masking process, both \textit{redaction} and \textit{replacement}, was successfully implemented. In redaction mode, sensitive entities such as names, dates, and ages were replaced with their respective entity type placeholder (e.g., "XXX-NAME", "XXX-DATE"). In replacement mode, realistic replacements were used for names and temporal entities from the list of full names and surnames extracted from the i2b2 2014 dataset, 
ensuring that the structure of the clinical document was maintained while still protecting the patient’s identity.

\begin{table}[t]
\centering
\scriptsize
\setlength{\tabcolsep}{3pt}
\begin{tabular}{p{0.9\columnwidth}}
\hline
\textbf{Replacement output} \\
\ttfamily
Oakley→Jones; 2065→2063; 3/67→01/65; 2068-12-05→2066-09-09; 37→34; 66→62 \\
\hline
\end{tabular}
\caption{Example replacements produced.}
\label{tab:replacement-processing-output}
\end{table}

\begin{figure}[t]
\centering
\includegraphics[width=0.4\textwidth]{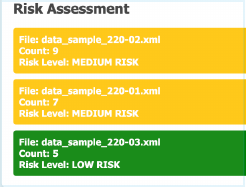}
    \caption{Risk Assessment Results for Batch upload of documents}
    \label{fig:risk-assess-result}
\end{figure}
\subsection{Risk Assessment Results}


Figure \ref{fig:risk-assess-result} risk assessment visualisation underscores the system's proficiency in identifying documents that still pose re-identification risks and provides actionable insights for further mitigating potential vulnerabilities in sensitive data.
The risk assessment results in Figure \ref{fig:risk-assess-result} effectively categorise documents based on the risk of re-identification. As shown, the system assigns each document a risk level — \textbf{high} (red), \textbf{medium} (yellow), or \textbf{low} (green) — depending on how many instances of unique entity contexts were found, referring to Section \ref{subsec:risk-assess} (Risk Assessment).

For example, \texttt{data\_sample\_220-02.xml} is marked with Medium Risk, having 9 unique contexts, while \texttt{data\_sample\_220-01.xml} similarly displays Medium Risk with 7 occurrences. These files flagged as medium risk suggest that while some sensitive information has been de-identified, there is still a non-negligible possibility of re-identification due to the unique contexts of certain entities.

In contrast, \texttt{data\_sample\_220-03.xml} is classified as Low Risk with only 5 instances of unique contexts, suggesting that most of the entities in this document share common contexts across the dataset, thus significantly lowering the chances of re-identification.


\subsection{Error Analysis and Limitations}
\begin{figure*}[t]
\centering
\includegraphics[width=0.95\textwidth]{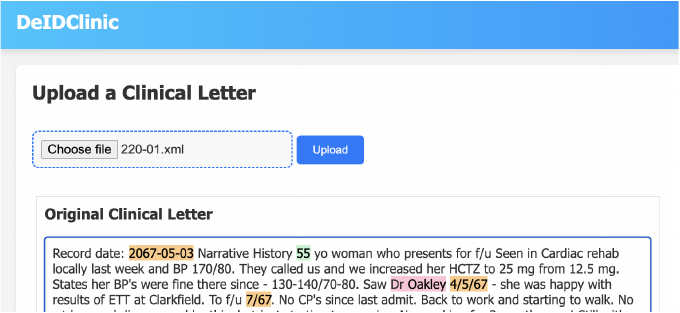}
    \caption{Boundary Error Example in MASK-BioBERT (\textbf{cardiac} domain text)}
    \label{fig:boundary-error}
\end{figure*}

Through manual inspection of the model outputs, we identified some primary sources of errors. First, \textit{boundary} errors occurred when the model slightly misidentified the start or end of an entity, a common issue in NER tasks, particularly for names that include prefixes or titles (e.g., ``Dr.~Oakley'' in Figure~\ref{fig:boundary-error}, Appendix). Second, the model produced \textit{false positives} for \textit{dates} and \textit{ages} by misclassifying numerical values unrelated to temporal information (e.g., medical measurements such as `2071' in Table~\ref{tab:sample-false-posi-bioBERT}), which negatively affected precision. Third, \textit{frequent terms} were occasionally over-masked as sensitive entities; for example, ``US'' was sometimes labeled as a location (Table~\ref{tab:sample-false-posi-bioBERT}), reducing precision by masking non-sensitive content.
Forth, overlapping entities and patterns introduced ambiguity, as the same text segment could be detected as multiple entity types by different methods (e.g., a place name misclassified as a person name). Fifth, the rule-based component occasionally fragmented multi-word entities into separate tokens, particularly for dates, where a single expression was split and treated as multiple independent entities.

From the experimental investigation outcomes of our work, several limitations and areas for improvement remain. First, the model was optimized for a single dataset due to the paucity of readily available data, resulting in dataset-specific performance and limited generalization to diverse clinical settings or real-world hospital deployments.
Second, due to limitations of computational facilities, we only tested on domain-specific BERT models for integration into the Mask framework, without using LLMs. \footnote{Recent work using LLMs for biomedical NER includes \cite{mazzucato2026advancements} for Dutch and Italian languages (preprint).}



\section{Conclusions and Future Work}
This work presents DeID-Clinic, a risk-aware pseudonymization framework for privacy-preserving processing of clinical free-text data. By integrating domain-adapted transformer models with document-level privacy risk assessment, the proposed system extends traditional de-identification pipelines beyond entity masking toward quantitative privacy evaluation.

Experimental results on the i2b2 dataset demonstrate strong performance in sensitive entity detection, while the proposed risk scoring framework enables identification of documents with elevated re-identification risk. This capability is particularly important in real-world privacy-sensitive applications, where automated de-identification alone may not fully eliminate privacy threats.
More broadly, this work highlights the importance of combining neural language models with explicit privacy risk modeling to support responsible data sharing. While evaluated on clinical data, the proposed framework is applicable to other privacy-critical domains, including legal and administrative text, aligning with emerging requirements for privacy-preserving language technologies.

Future work will focus on 
extending the proposed framework in several directions: 1) evaluate the risk assessment module across multiple datasets and domains to further validate its effectiveness in estimating re-identification risk; 2)  integrate newer LLMs, which may improve performance on context-dependent entity types such as professions and organizations; 3) explore adaptive risk thresholds and user-configurable privacy settings to support more flexible deployment in real-world privacy-sensitive environments.



\section{Ethical Statement}
The data we used for this work is already de-identified and anonymized by the shared task organisers who released the data for research purposes only. We did not use any third party commercial platforms to disclose the data. 

\section{Acknowledgement}
We are grateful to the reviewers for valuable comments.
Funded by the European Union under Horizon Europe Work Programme 101057332, views and opinions expressed are however those of the author(s) only and do not necessarily reflect those of the European Union or the European Health and Digital Executive Agency (HaDEA). Neither the European Union nor the granting authority can be held responsible for them.
The UK team are funded under the Innovate UK Horizon Europe Guarantee Programme, UKRI Reference Number: 10041120.
WDP and GN are grateful for the support from the grant “Assembling the Data Jigsaw: Powering Robust Research on the Causes, Determinants and Outcomes of MSK Disease”, and the grant “Integrating hospital outpatient letters into the healthcare data space” (EP/V047949/1; funder: UKRI/EPSRC).

\section{Bibliographical References}\label{sec:reference}

\bibliographystyle{lrec2026-natbib}
\bibliography{lrec2026-example}


\bibliographystylelanguageresource{lrec2026-natbib}
\bibliographylanguageresource{languageresource}

\clearpage
\section{Appendix}
\section*{Data Statistics and Training}

\begin{figure}[h]
\centering
\includegraphics[width=0.45\textwidth]{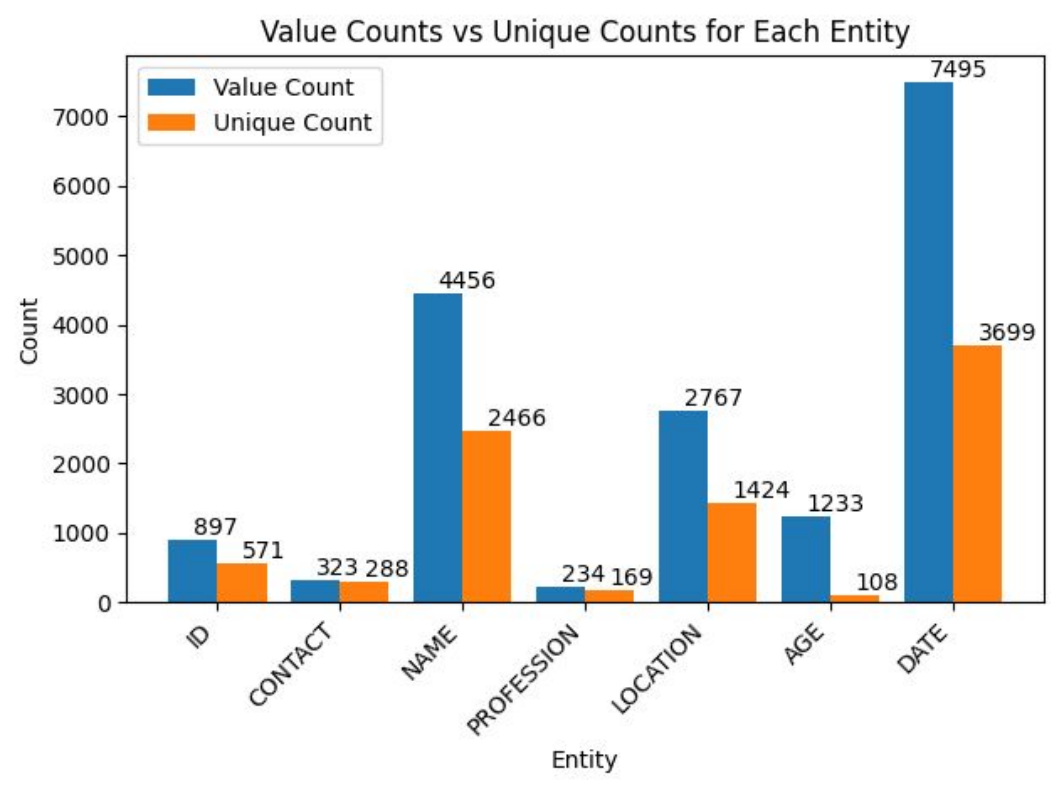}
    \caption{Entity occurrence distribution in the i2b2 dataset, showing value counts and unique counts for each entity type
    }
    \label{fig:entity-occur}
\end{figure}

\begin{figure}[h]
\centering
\includegraphics[width=0.45\textwidth]{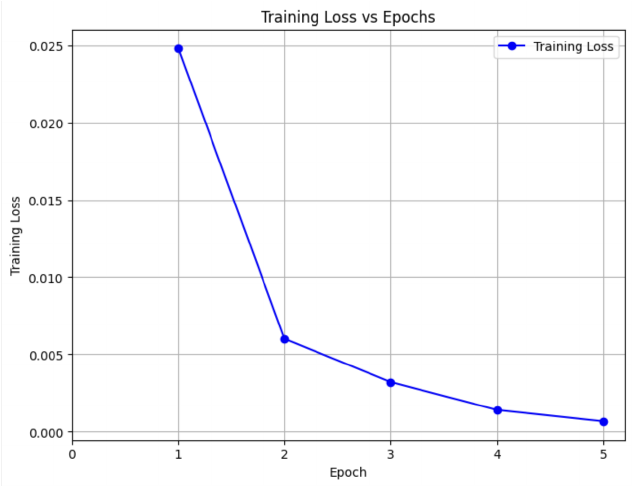}
    \caption{Training curve visualisation}
    \label{fig:train-curve-bioBERT}
\end{figure}

Figure \ref{fig:entity-occur} shows the entity occurrence distribution in the i2b2 dataset.
Figure \ref{fig:train-curve-bioBERT} is an example of how loss evolved during training, which is indicative of the model's learning trajectory.

\begin{figure*}[t]
\centering
\includegraphics[width=0.95\textwidth]{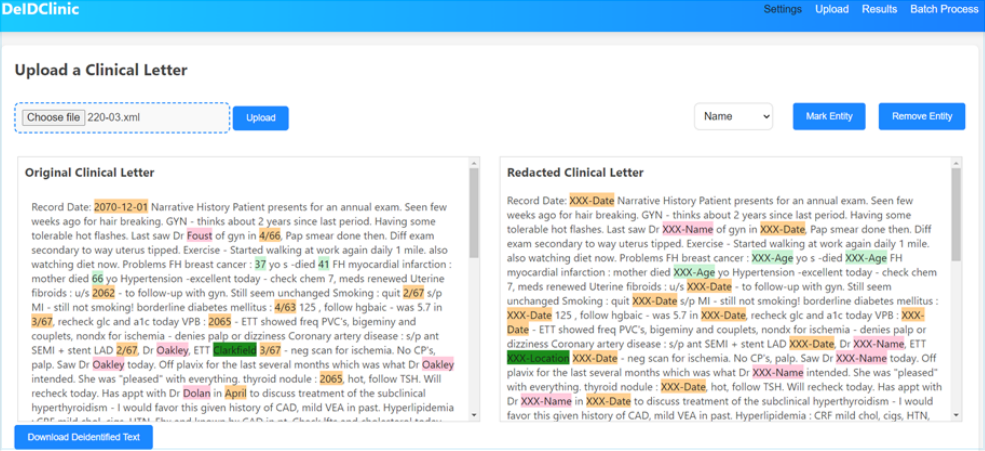}
    \caption{\textbf{Interface Demo} with De-identification output using uploaded letter (\textbf{cancer} domain text)}
    \label{fig:output-deid}
\end{figure*}

\section*{Tools and Technologies in Detail}
Some technical details are listed below:
\begin{itemize}
    \item This work is built on the MASK API developed by the Northern Care Alliance NHS Foundation Trust, available publicly on GitHub \cite{milosevic2020mask,NCA2021}.
    \item Risk Assessment Metrics: Libraries such as Scikit-learn and SciPy were used for calculating re-identification risk metrics \cite{Dwork2006}.
    \item Google Colab provides the necessary T4-GPU resources for fine-tuning.


\end{itemize}

\section*{Platform Interface: Human-in-the-loop}\label{sec:app:interface}

The de-identification platform, as demonstrated in Figure \ref{fig:output-deid}, can support both single file and multiple files processing. The De-identificaiton procedure is:

\begin{itemize}
    \item Load Finetuned (saved) models, e.g. MASK-BioBERT/ClinicalBERT
    \item Run De-identification using the loaded model
    \item Mark/Remove Entities Option, human-in-the-loop, editable results
    \item Store/Download final output
\end{itemize}

\end{document}